\documentclass[a4paper]{article}
\usepackage{authblk}
\usepackage{balance}
\usepackage{spconf,amsmath,graphicx}
\usepackage{graphicx}
\usepackage{pdfpages}
\usepackage{cite}

\usepackage{subfig} 
\usepackage{algorithm}
\usepackage{algpseudocode}
\usepackage[T1]{fontenc} 
\usepackage{array}
\usepackage{url}
\usepackage{float}

\title{Systematic study of color spaces and components\\for the segmentation of sky/cloud images}

\twoauthors
  {Soumyabrata Dev, Yee Hui Lee \thanks{This research is funded by the Defence Science and Technology Agency (DSTA), Singapore.}}%
	{School of Electrical and Electronic Engineering\\
    Nanyang Technological University (NTU)\\
    Singapore 639798}
  {\hspace{-10mm}Stefan Winkler}
	{\hspace{-10mm}Advanced Digital Sciences Center (ADSC)\\
   \hspace{-10mm}University of Illinois at Urbana-Champaign\\
   \hspace{-10mm}Singapore 138632}

\begin{document}
\algnewcommand\algorithmicswitch{\textbf{switch}}
\algnewcommand\algorithmiccase{\textbf{case}}
\algnewcommand\algorithmicassert{\texttt{assert}}
\algnewcommand\Assert[1]{\State \algorithmicassert(#1)}%
\algdef{SE}[SWITCH]{Switch}{EndSwitch}[1]{\algorithmicswitch\ #1\ \algorithmicdo}{\algorithmicend\ \algorithmicswitch}%
\algdef{SE}[CASE]{Case}{EndCase}[1]{\algorithmiccase\ #1}{\algorithmicend\ \algorithmiccase}%
\algtext*{EndSwitch}%
\algtext*{EndCase}%

\maketitle

\begin{abstract}
Sky/cloud imaging using ground-based Whole Sky Imagers (WSI) is a cost-effective means to understanding cloud cover and weather patterns. The accurate segmentation of clouds in these images is a challenging task, as clouds do not possess any clear structure. Several algorithms using different color models have been proposed in the literature. This paper presents a systematic approach for the selection of color spaces and components for optimal segmentation of sky/cloud images. Using mainly principal component analysis (PCA) and fuzzy clustering for evaluation, we identify the most suitable color components for this task.
\end{abstract}


\section{Introduction}
\label{sec:intro}
Cloud analysis plays an important role in weather prediction, climate modeling, solar irradiation measurement for renewable energy generation, and analysis of signal attenuation in satellite and other space-to-ground communications \cite{site_diversity}. The conventional manner of performing such cloud analysis via geo-stationary satellite images does not provide sufficient resolution for some of these applications.  Ground-based Whole Sky Imagers (WSIs) can provide higher spatial and temporal resolution for highly-localized cloud analysis, and several types have been developed \cite{WSI_UCSD,Souza,Long,WAHRSIS}.

The detection of clouds from sky images is challenging as clouds do not possess any definite structure, contour, shape, or size. As a result, color has been used as the predominant feature for sky/cloud segmentation. Numerous techniques based on different color models and spectral wavelengths have been proposed in the literature to solve this problem. However, the selection of color models and channels in the existing algorithms seem ad-hoc in manner, without systematic analysis or comparison.  This paper aims to address this issue by providing a structured review and evaluation of color models for the segmentation of sky/cloud images.  For this purpose, we make use of color distribution characteristics, principal component analysis (PCA), and clustering methods.\footnote{~The source code of all simulations in this paper is available online at \url{https://github.com/Soumyabrata/color-channels}.}  For our evaluation, we use the HYTA image database \cite{Li2011}, which contains 32 sky/cloud images with segmentation ground truth.  Our comprehensive analysis is able to clearly identify the most suitable color models and components for the analysis of cloud/sky images.

The paper is structured as follows. Section \ref{sec:approach} describes our approach, including the color models, techniques, tools, and database used for the subsequent evaluation.  Section \ref{sec:evaluation} presents the results of the quantitative performance comparison. Section \ref{sec:conclusions} concludes the paper.

\section{Approach \& Tools}
\label{sec:approach}
\subsection{Color Models}
\label{sec:colorchannels}
In the literature, several techniques based on different color models have been proposed for sky/cloud segmentation. Long et al.\  \cite{Long} use the ratio of red and blue channels ($R/B$) to detect clouds using appropriate threshold values. Calb{\'{o}} and Sabburg \cite{Calbo2008} use the same ($R/B$) ratio to derive statistical features (mean, standard deviation, entropy etc.) of the clouds and subsequently classify the sky/cloud images into different cloud types. Heinle et al.\ \cite{Heinle2010} utilize a k-nearest-neighbor classifier using the difference of red and blue channels ($R-B$) to classify cloud types. Souza-Echer et al.\ \cite{Souza} choose saturation for the estimation of cloud coverage. Mantelli-Neto et al.\ \cite{Sylvio} classify clouds by exploiting the locus of pixels in the RGB color model. Most recently, Li et al.\ use a normalized difference of blue and red channels ($\frac{B-R}{B+R}$) \cite{Li2011} for cloud detection.

The choices of color models employed in existing papers are based on empirical observations about the color distributions of cloud and sky pixels.  However, there exists no systematic analysis to determine the most suitable color channel(s) for this purpose. The purpose of this paper is to present a unified approach in the analysis of sky/cloud images using different color channels and to identify the color channels with the best results.

We consider the following color spaces and components for analysis (see Table~\ref{ps}): RGB, HSV, YIQ, CIE $\mbox{L}^{*}\mbox{a}^{*}\mbox{b}^{*}$, three forms of red-blue combinations ($R/B$, $R-B$, $\frac{B-R}{B+R}$), and chroma $C=\mbox{max}(R,G,B)-\mbox{min}(R,G,B)$. In addition to the color channels $c_{1-9}$ and $c_{13-16}$ used in the existing literature, we also include $c_{10-12}$ and $c_{16}$, as separating chromatic and achromatic information may prove beneficial for sky/cloud image segmentation.

\begin{table}[t]
\small
\centering
\setlength{\tabcolsep}{4pt} 
\begin{tabular}{c|c||c|c||c|c||c|c||c|c||c|c}
  \hline
  $c_{1}$ & R & $c_{4}$ & H & $c_{7}$ & Y & $c_{10}$ & $L^{*}$ & $c_{13}$ & $R/B$ & $c_{16}$ & $C$\\
  $c_{2}$ & G & $c_{5}$ & S & $c_{8}$ & I & $c_{11}$ & $a^{*}$ & $c_{14}$ & $R-B$& $ $ & $ $\\
  $c_{3}$ & B & $c_{6}$ & V & $c_{9}$ & Q & $c_{12}$ & $b^{*}$ & $c_{15}$ & $\frac{B-R}{B+R}$ & $ $ & $ $\\
  \hline
\end{tabular}
\caption{Color spaces and components used for analysis.}
\label{ps}
\end{table}

\subsection{Color Distribution}
\label{S:PBI}
Since we are essentially trying to distinguish between two classes of pixels (sky and clouds), a color model with a bimodal distribution can facilitate this task.  Color components with a high degree of bimodality are not only good candidates for determining the number of clusters in a sample image, but can also be used to decide whether or not the sample image requires segmentation in the first place. Pearson's Bimodality Index (PBI) is a popular statistic to evaluate the bimodal behavior quantitatively \cite{pearson}. It is defined as:
\begin{equation}
\label{eq:pbi_defn}
\mbox{PBI}= b_2-b_1,
\end{equation}
where $b_{2}$ is the kurtosis and $b_{1}$ is the square of skewness.  A PBI value close to $1$ indicates highly bimodal distributions.

\subsection{Principal Component Analysis}
\label{S:PCA}
We use the Principal Component Analysis (PCA) to (a) check the correlation and similarity between different color components, and (b) determine those color components that capture the greatest variance.  

The PCA is computed as follows.  Let us assume a sample image $\textbf{X}_{i}$ of dimension $m \times n$ pixels from a set of $N$ images ($i=1,...,N$).  Each color channel of the sample image is reshaped into a straight vector $\textbf{c}_{j}$ of dimensions $mn \times 1$. These column vectors are stacked alongside each other to form a matrix $\hat{\textbf{X}_{i}}$ of dimensions $mn \times 16$:
\begin{equation}
\label{eq:eq1}
\hat{\textbf{X}_{i}}=[\textbf{c}_{1}, \textbf{c}_{2},..,\textbf{c}_{j}..,\textbf{c}_{16}] ; j=1,...,16
\end{equation}

The ranges of these 16 color channels are different and need to be normalized so that no color channel is under- or over-represented in the PCA analysis. Each of the 16 color channels is normalized using its mean and standard deviation across all the images in the dataset, thereby generating the new image representation $\ddot{\textbf{X}_{i}}$ with zero mean and unit variance:
\begin{equation}
\label{eq:eq3}
\ddot{\textbf{X}_{i}}= [\frac{\textbf{c}_{1}-\bar{c_{1}}}{\sigma_{c_{1}}}, \frac{\textbf{c}_{2}-\bar{c_{2}}}{\sigma_{c_{2}}},..,\frac{\textbf{c}_{j}-\bar{c_{j}}}{\sigma_{c_{j}}},..,\frac{\textbf{c}_{16}-\bar{c_{16}}}{\sigma_{c_{16}}}]
\end{equation}

Subsequently the covariance matrix $\textbf{M}_{i}$ is computed for each of the $\ddot{\textbf{X}_{i}}$. Let the eigenvector $\textbf{e}_{ij}$ and eigenvalue $\lambda_{ij}$ ($j=1,...,16$) be obtained from the eigenvalue decomposition of the matrix $\textbf{M}_{i}$. For the $i^\textrm{th}$ image, the eigenvectors $\textbf{e}_{ij}$ corresponding to the largest eigenvalues $\lambda_{ij}$ encompass an orthonormal vector subspace. The eigenvectors and eigenvalues are calculated to project the multi-dimensional vector space into lower dimensions and to reveal the underlying relationship amongst them. 

\subsection{Clustering}
\label{sec:clustering}
For clustering the sky/cloud images, we employ the fuzzy c-means algorithm \cite{Bezdek,Jawahar} to assign probabilities of cloud detection to the set of pixels of the input image. The algorithm for the effective segmentation of clouds from the sky/cloud images employs the minimization of the following objective function:
\begin{equation}
\label{eq:eq4}
J= \sum\limits_{r=1}^{2} \sum\limits_{s=1}^{mn} p_{r}(\textbf{x}_{s})^{\tau}d(\textbf{x}_{s},\textbf{v}_{r}),
\end{equation}
where $\tau$ is called the fuzziness index, which controls the degree of fuzziness during the clustering process; we set $\tau=2$.  $d(\textbf{x}_{s},\textbf{v}_{r})$ denotes the 2D Euclidean norm between the input vector $\textbf{x}_{s}$ and the cluster centers $\textbf{v}_{r}$. Both $\textbf{v}_{1}$ and $\textbf{v}_{2}$ are vectors of dimension $k$, where $k$ can take any positive integer number.

\section{Evaluation \& Results}
\label{sec:evaluation}

\subsection{Sky/Cloud Image Database}
\begin{figure}[htb]
\centering
\includegraphics[height=0.64in]{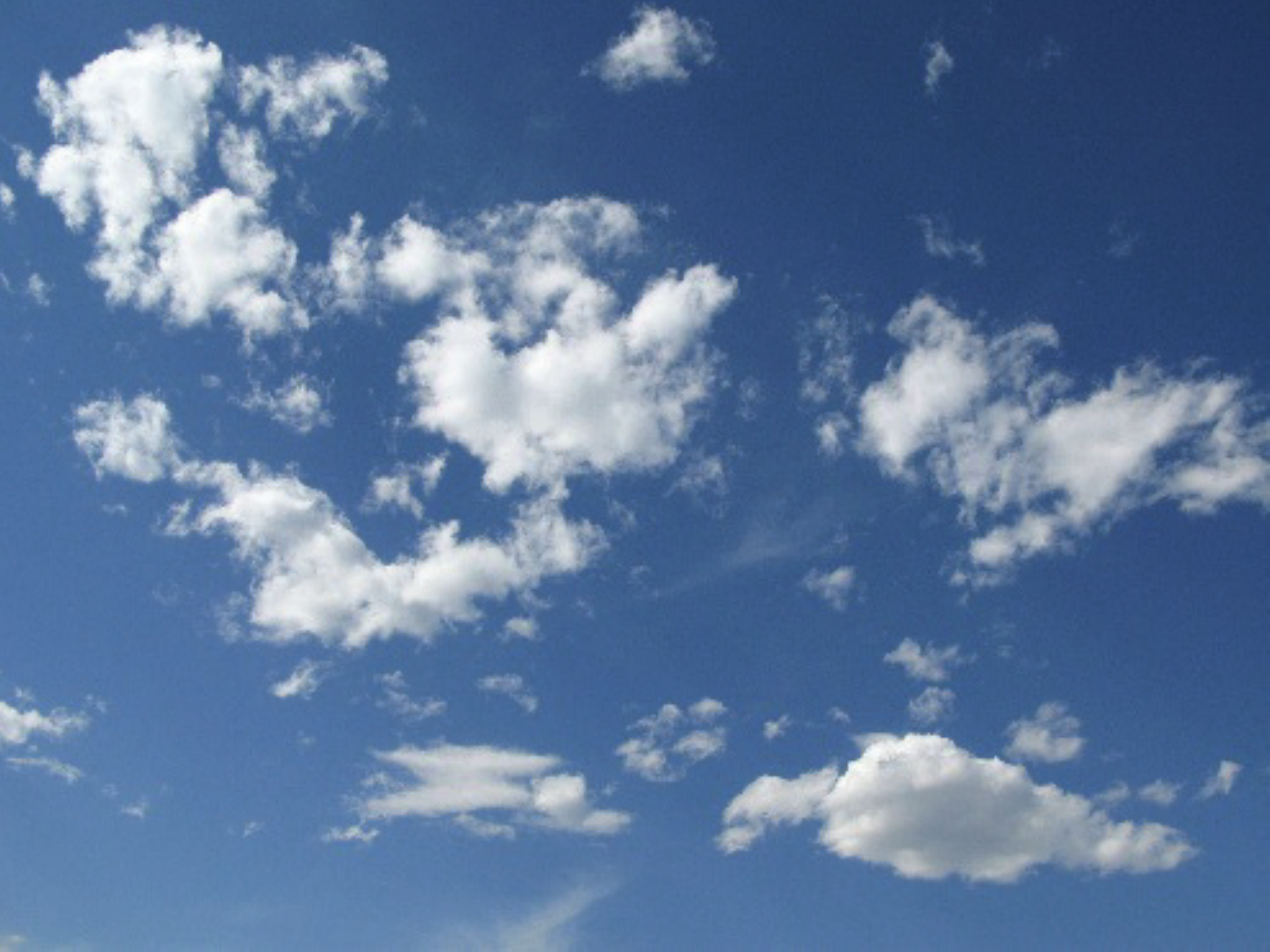}\hspace{0.5mm}
\includegraphics[height=0.64in]{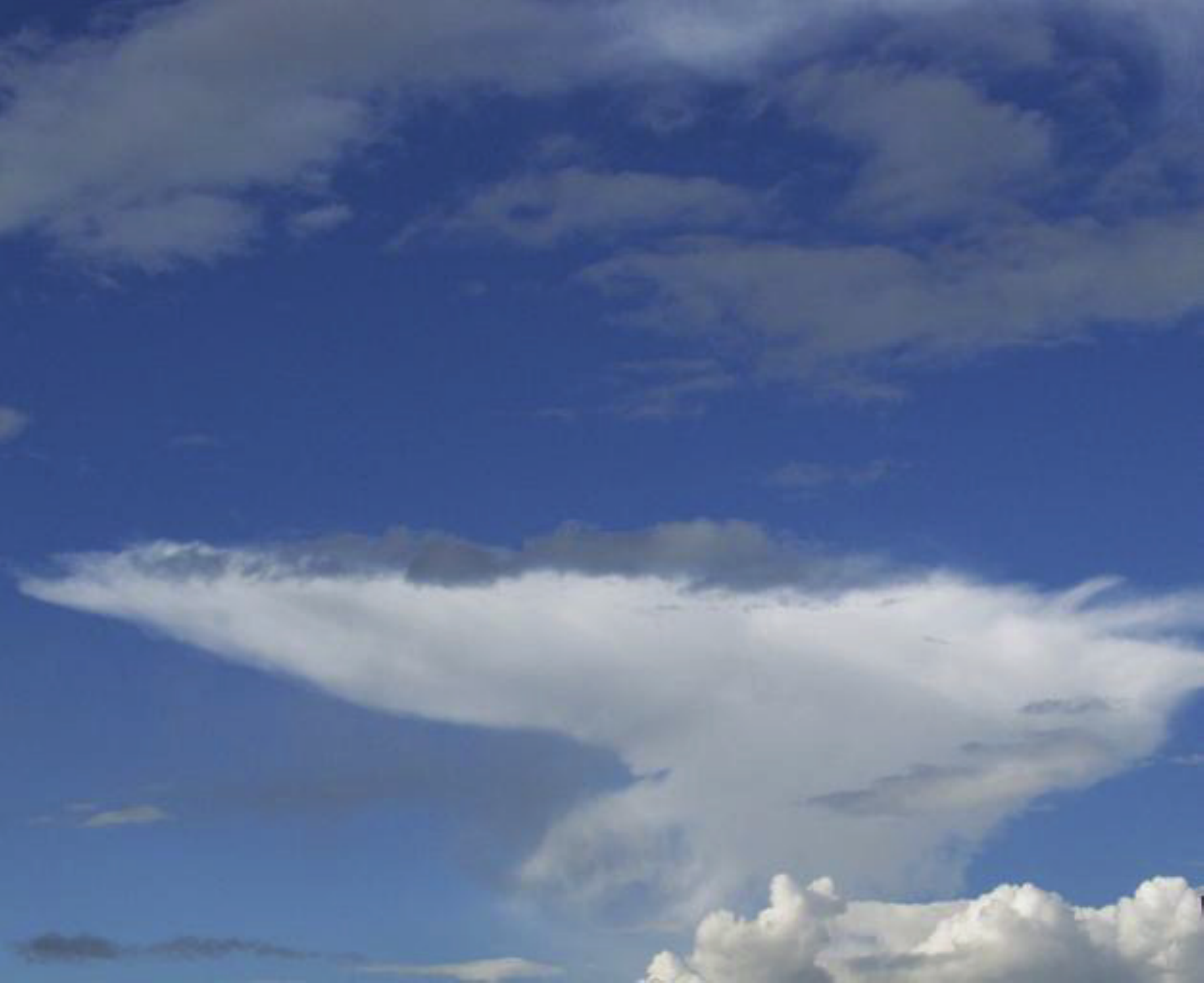}\hspace{0.5mm}
\includegraphics[height=0.64in]{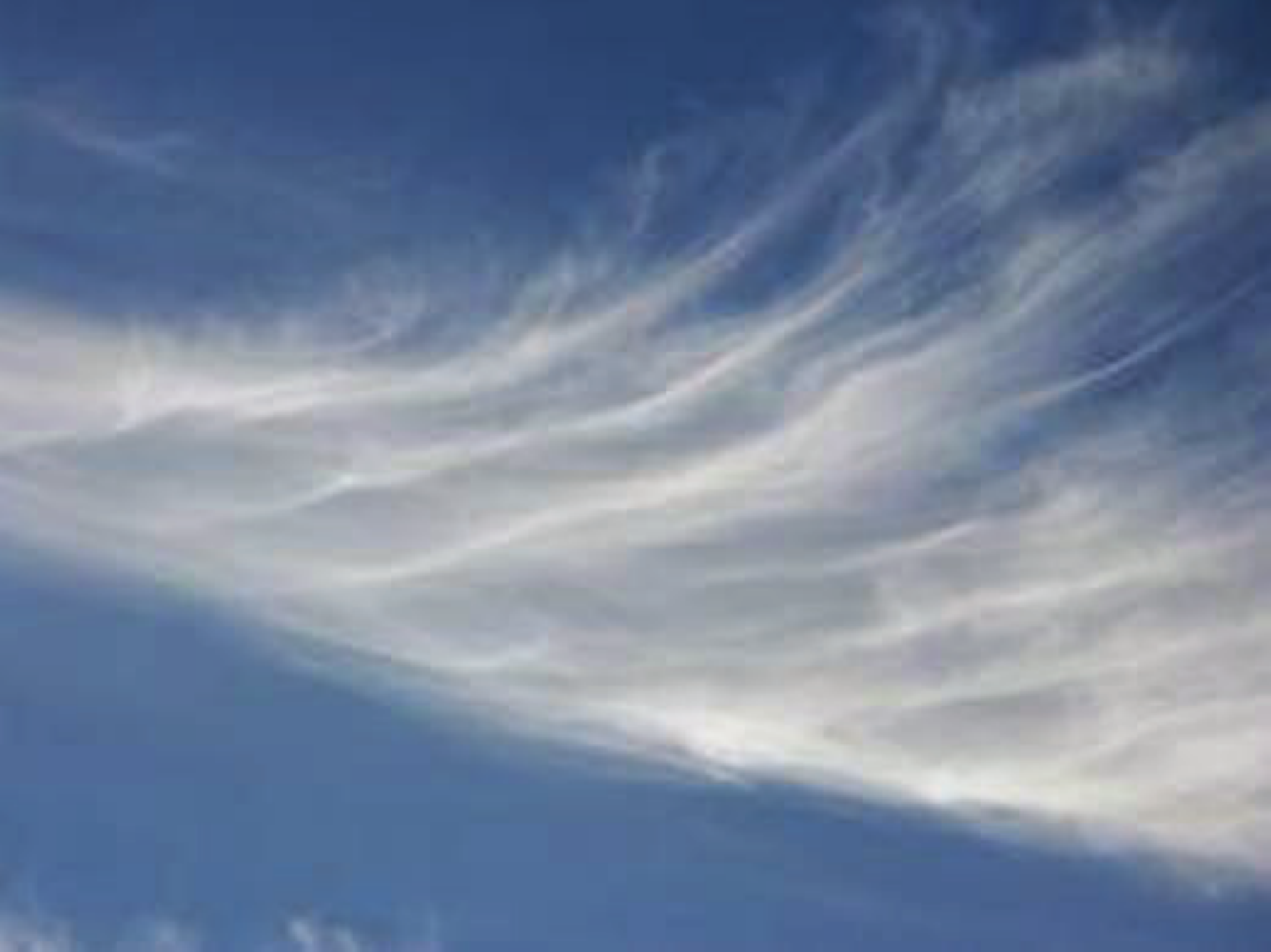}\hspace{0.5mm}
\includegraphics[height=0.64in]{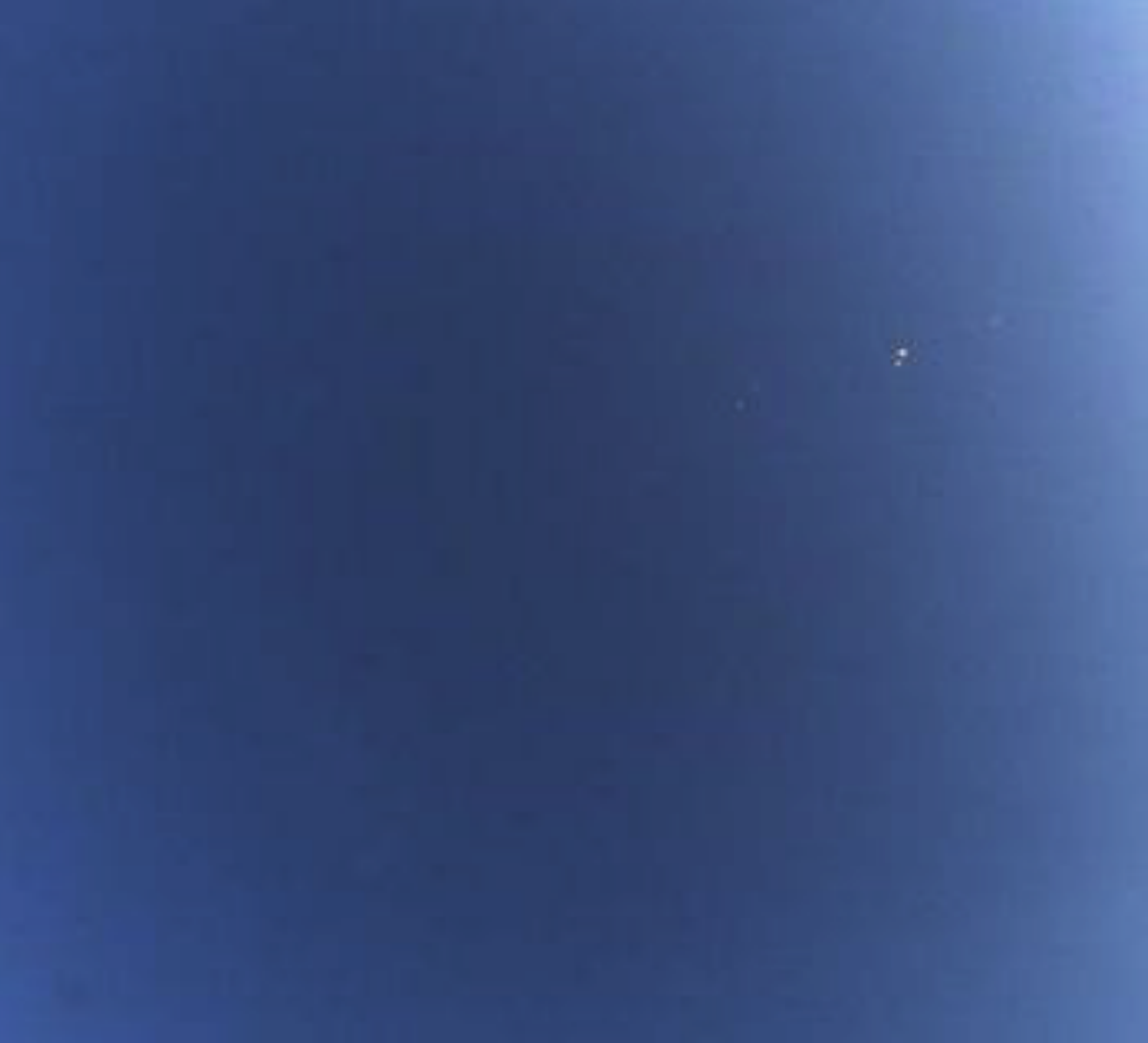}\\\vspace{1mm}
\includegraphics[height=0.64in]{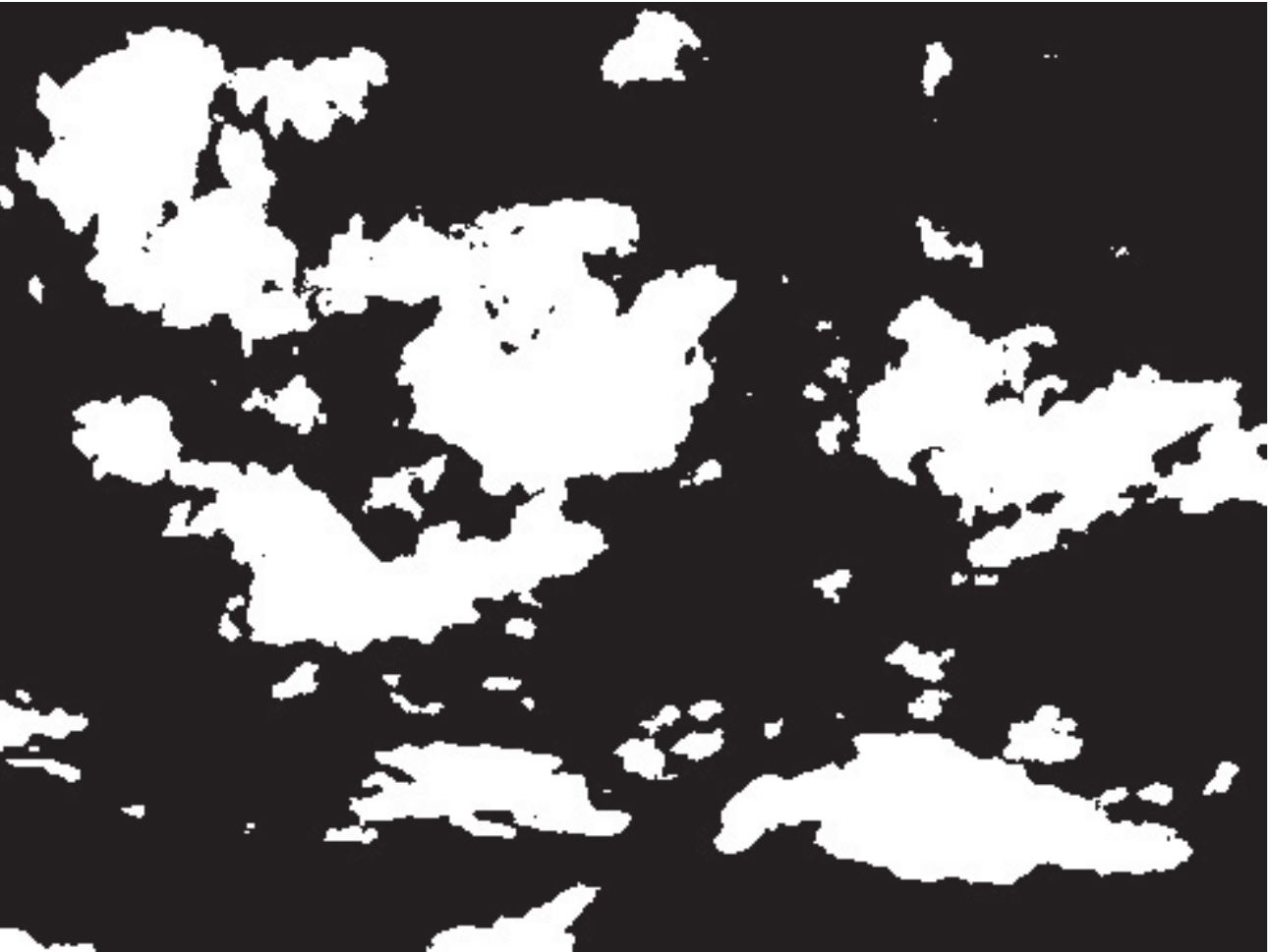}\hspace{0.5mm}
\includegraphics[height=0.64in]{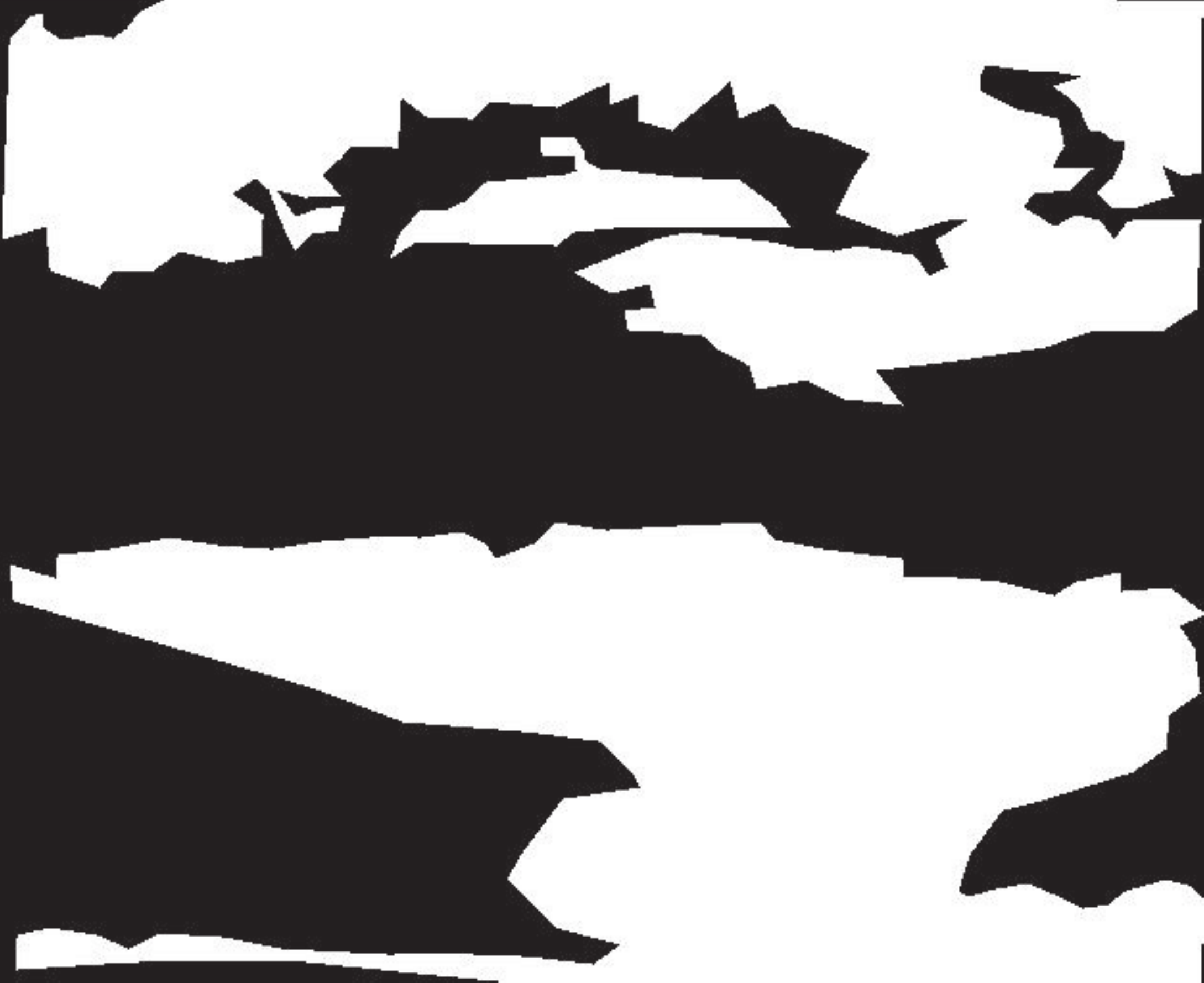}\hspace{0.5mm}
\includegraphics[height=0.64in]{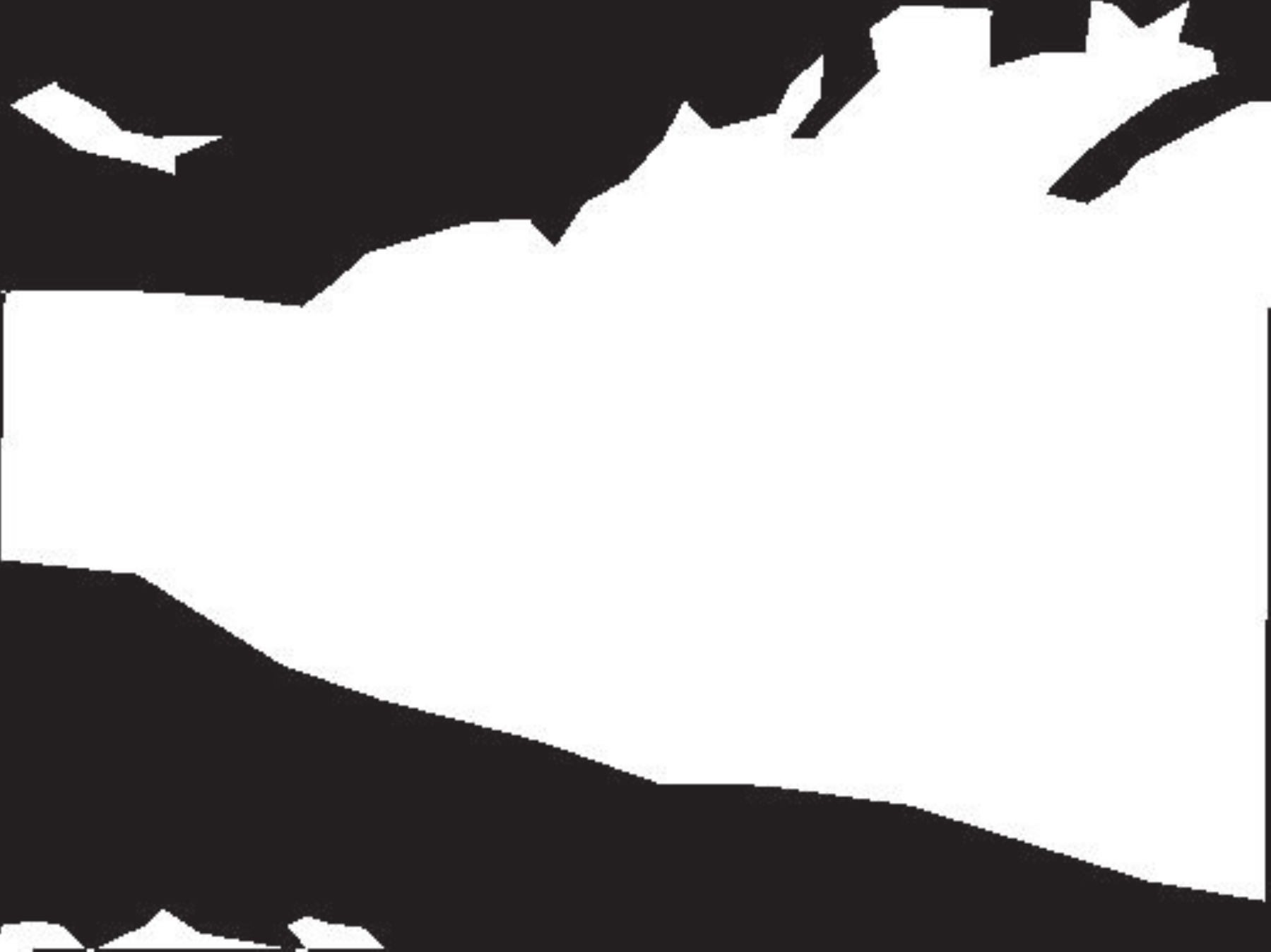}\hspace{0.5mm}
\includegraphics[height=0.64in]{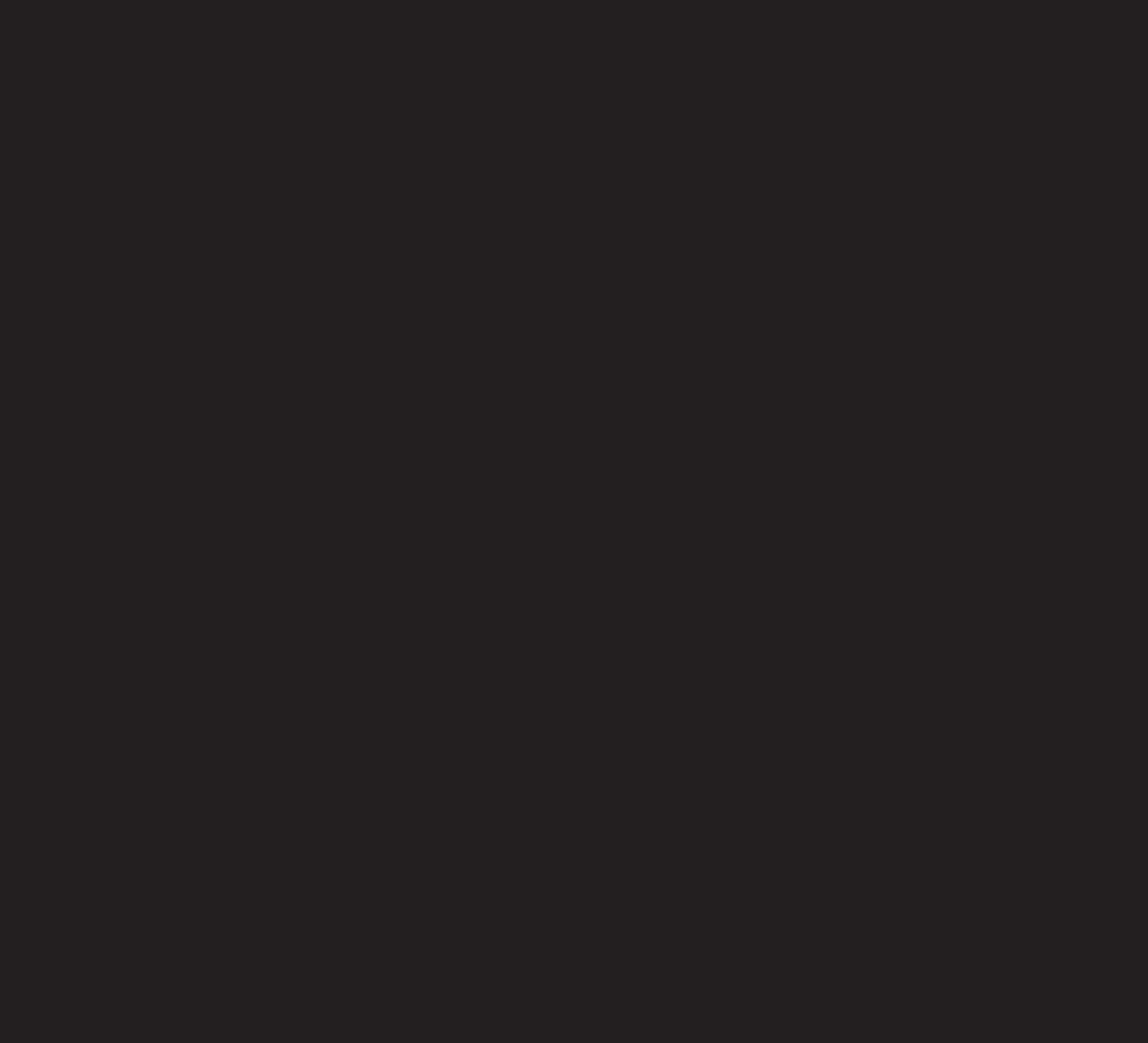}
\caption{Sample images (top row) along with corresponding sky/cloud segmentation ground truth (bottom row) from the HYTA database.}
\label{fig:sample_HYTA}
\end{figure}

To our knowledge, the only currently available database for sky/cloud images with segmentation ground truth is the HYTA database \cite{Li2011}. It consists of 32 distinct images of various sky/cloud conditions. Fig.~\ref{fig:sample_HYTA} shows a few sample images of the HYTA database along with their corresponding binary ground truth. 

\subsection{Distribution Bimodality}
The segmentation of input image into two classes (sky and clouds) becomes easier for those color channels which exhibit higher bimodality. The bimodal behavior of a color channel for the concatenated distribution is measured using Pearson's Bimodality Index (PBI) as described in Section \ref{S:PBI}.  The results are summarized in Table~\ref{PBI16}. Out of the 16 color channels, $c_{5}$ (S), $c_{1}$ (R), and $c_{13}$ ($R/B$) have the lowest PBI and thus exhibit the most bimodal distributions, indicating that the segmentation of images into two distinct classes should work well in these color channels.

\begin{table}[htb]
\small
\centering
\setlength{\tabcolsep}{3pt} 
\begin{tabular}{c|c||c|c||c|c||c|c||c|c||c|c}
  \hline
  $c_{1}$ & \textbf{2.24} & $c_{4}$ & 3.11 & $c_{7}$ & 2.71 & $c_{10}$ & 2.86 & $c_{13}$ & \textbf{2.27} & $c_{16}$ & 4.27\\
  $c_{2}$ & 2.83 & $c_{5}$ & \textbf{1.94} & $c_{8}$ & 3.98 & $c_{11}$ & 8.85 & $c_{14}$ & 4.43 & $ $ & $ $\\
  $c_{3}$ & 3.25 & $c_{6}$ & 3.26 & $c_{9}$ & 5.96 & $c_{12}$ & 4.59 & $c_{15}$ & 2.92 & $ $ & $ $\\
  \hline
\end{tabular}
\caption{PBI values for all color channels. The most bimodal channels are highlighted in bold.}
\label{PBI16}
\end{table}

\subsection{PCA}
\label{sec:winners}

\begin{figure}[htb]
\hspace{-0.1in}\includegraphics[width=1.1\columnwidth]{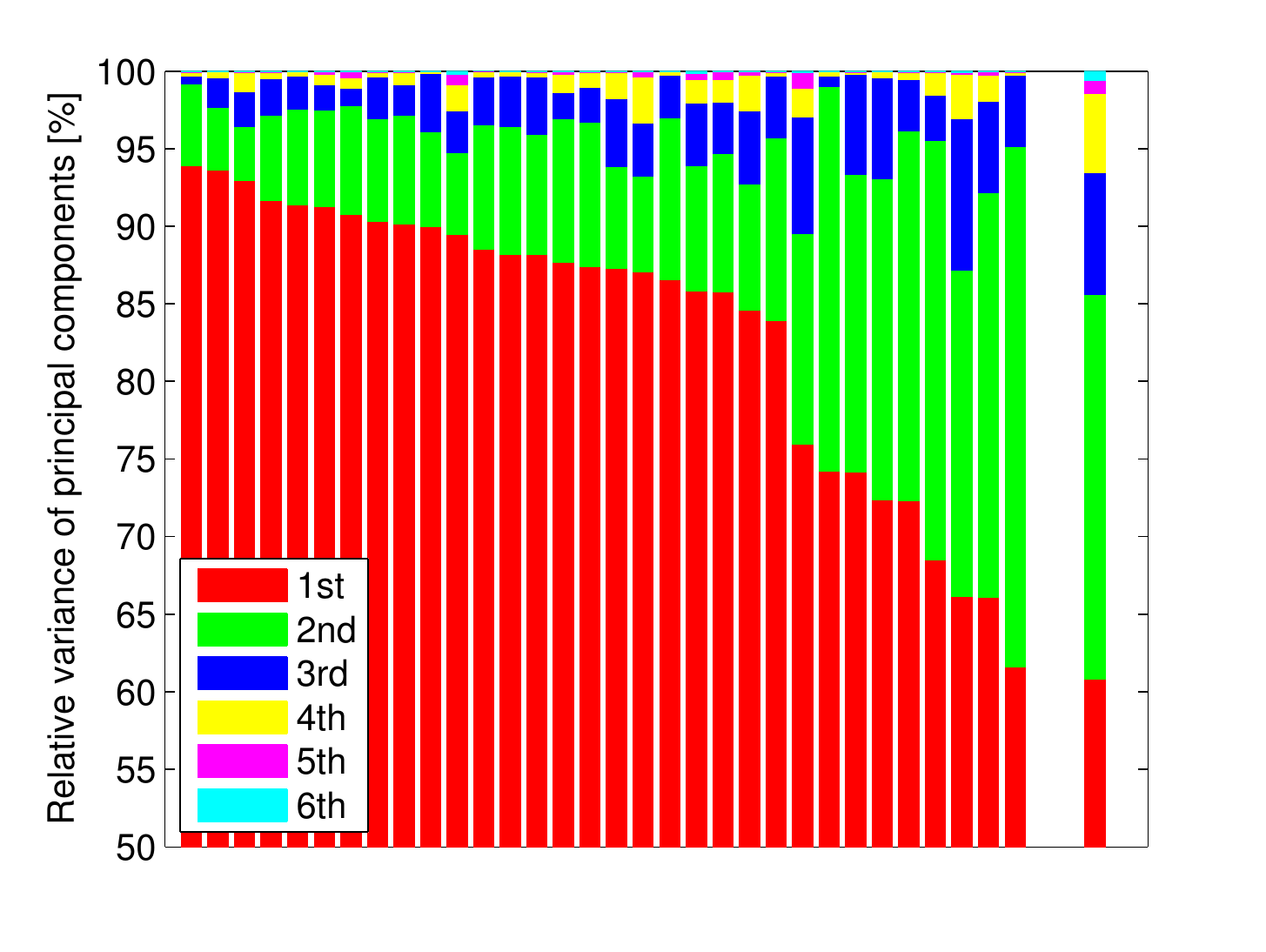}
 \caption{Distribution of the variance across the principal components for all images of the HYTA database. The separate bar on the right shows the variance distribution for the concatenation of all images of the entire database.
\label{fig:pca16_stefan}}
\end{figure}

The distribution of the amount of variance captured by the principal components is shown in Fig.~\ref{fig:pca16_stefan}. The combination of first and second principal components capture a large majority ($95.4$\%) of variance in the input data across the individual images, and over $90$\% for the entire database. 

In order to find the most significant vector for the detection of clouds, the contributing effect of the individual variables on the primary principal component axis is analyzed. Every  color channel has different loading factors on the most important principal component axis, and consequently each of them has a different contribution to the common variance captured in the corresponding data matrix $\ddot{\textbf{X}_{i}}$. These individual loading factors are essentially the projections of the input vectors (color channels) on the principal components. The significant color channels in terms of relative contribution to the first principal component axis are $c_{13}$ ($R/B$), $c_{15}$ ($\frac{B-R}{B+R}$, and $c_{5}$ ($S$).

Extending our analysis to finding the most significant pair of color channels, we need to understand a pair's cumulative contribution to the common variance. The sum of the squares of the corresponding loading factors for different color channels in the concatenated distribution represents the cumulative common variance captured by the pair.  The pairs that have the highest cumulative loading on the first principal component are $c_{13}$-$c_{15}$, $c_{5}$-$c_{13}$, and $c_{5}$-$c_{15}$.

\begin{figure}[htb]
\hspace{-0.24in}\includegraphics[width=0.58\textwidth]{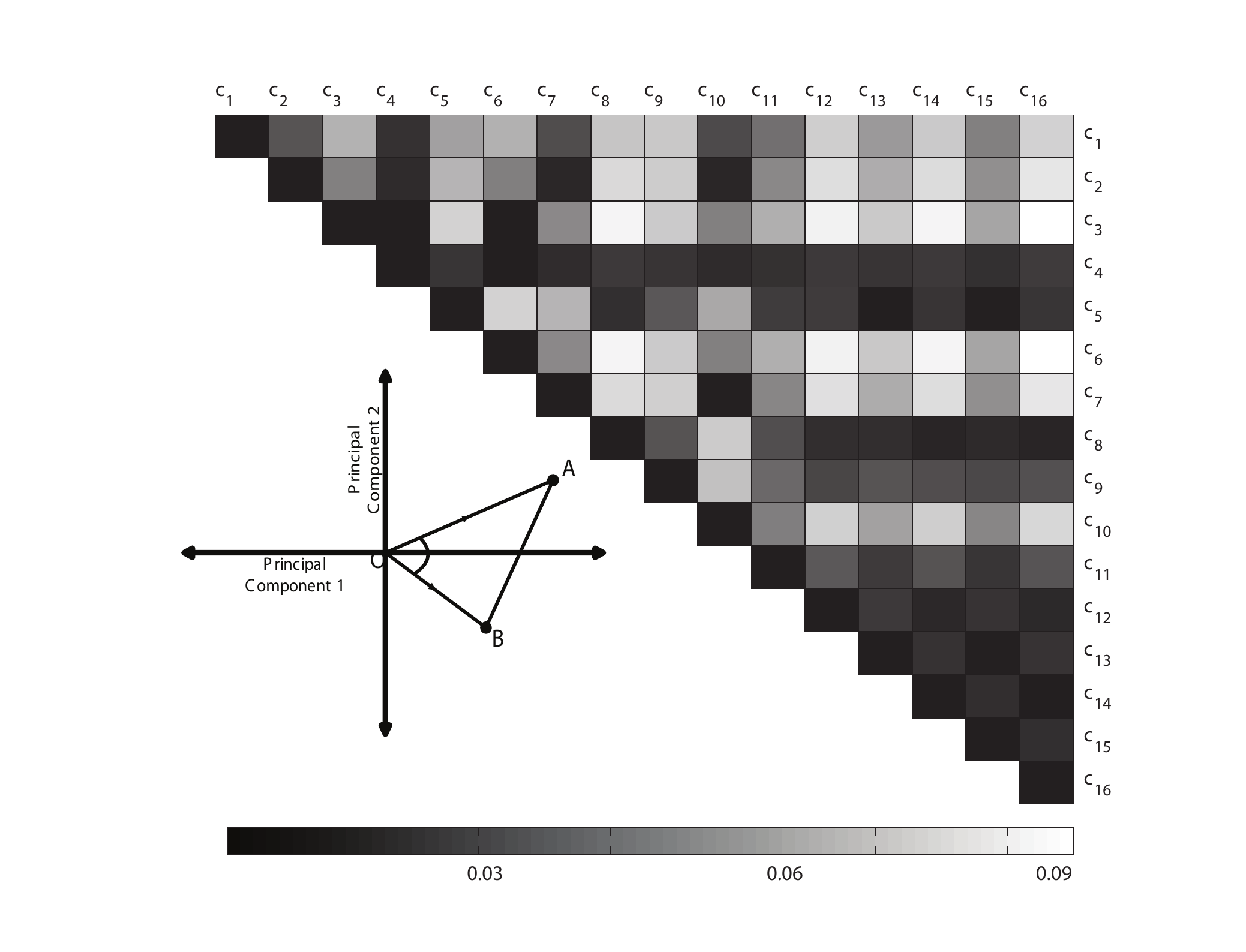}
\caption{Distribution of area \emph{AOB} for all input vector pairs, where \emph{OA} and \emph{OB} are two color channels. The brightness of each square represents the area captured by the corresponding pair.}
\label{fig:area_grid_v7}
\end{figure}

Another aspect to consider is the orthogonality between the individual color channels. For this purpose, we compute the triangular area enclosed by two given input vectors (color channels) projected on the plane spanned by the first and second principal components. There are ${{16}\choose{2}}=120$ combinations of two color channels. The triangular area for all unique cases is shown in Fig.~\ref{fig:area_grid_v7}, where the brightness of each square represents the area captured by the corresponding pair.  As expected, certain color channel pairs (the dark patches) are highly correlated, for example the green channel ($c_2$) with most luma channels ($c_7$, $c_{10}$), the various $R$-$B$ combinations ($c_{13-15}$) with one another, and so on.

\subsection{Clustering}
\label{sec:results}
In order to verify the efficacy of specific color channels in the detection of cloud pixels, 1D and 2D clustering results are presented across all 16 color channels for all the images of HYTA and evaluated using precision, recall, and F-score. Since the ground truth provided by the HYTA database are binary images consisting of two classes, namely sky and cloud, we convert the probabilistic distribution of cloud pixels $p_{r}(\textbf{x}_{s})$ (cf.\ Section~\ref{sec:clustering}) to a binary image by thresholding, i.e.\ a pixel having a probability of cloud $>50\%$ is assigned to a cloud pixel and others to sky pixels.

\begin{figure}[H]
\vspace{-3mm}
\hspace{-0.5in}\includegraphics[width=0.6\textwidth]{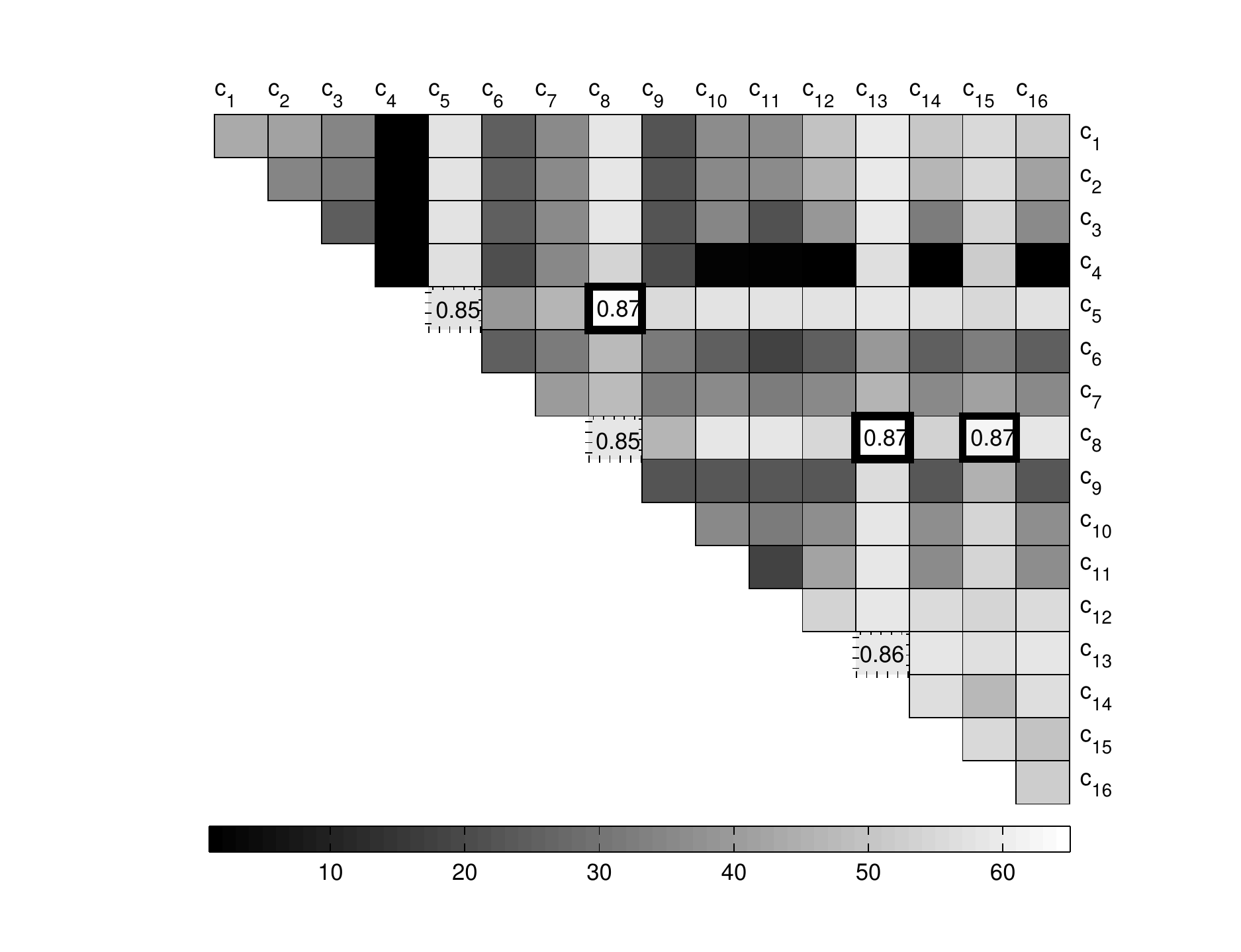}
 \caption{F-Scores for all unique cases of 1D (along the diagonal) and 2D clustering, represented on a brightness scale.  The highest F-Scores for 1D (dotted frames) and 2D cases (bold frames) are indicated numerically.
\label{fig:2d_result}}
\end{figure}

There are 16 color channels for 1D clustering.  In a similar manner, there are ${{16}\choose{2}}$ unique color channel combinations for 2D-clustering. Their performance is illustrated in Fig.~\ref{fig:2d_result} in an upper-triangular grid, where the intensity of a particular grid denotes the F-score of that particular color channel pair; the diagonal elements represent the F-scores of the corresponding 1D color channels.  The most promising color channels for 1D-clustering are $c_{13}$ ($R/B$), $c_{15}$ ($\frac{B-R}{B+R}$), and $c_{5}$ ($S$) with high precision, recall, and F-scores.  For 2D, the best combinations include $c_5$-$c_8$, $c_8$-$c_{13}$, and $c_8$-$c_{15}$ ($c_8$ representing essentially another variety of red-blue difference), with several other close contenders.

\subsection{Discussion}
\label{sec:relationship}

The clustering results as illustrated in Section \ref{sec:results} confirm the efficacy of the color channels determined earlier in Section \ref{sec:winners}. Color channels $c_{5} $($S$), $c_{1}$ ($R$), and $c_{13}$ ($R/B$), which exhibit a highly bimodal behavior, are relatively good indicators of clustering performance.  We also found the loading factors of the first eigenvector (i.e.\ the re-projection of the data points on the first principal component) to be a reasonable guide for the choice of good color models; the goodness of fit between loading factors and corresponding F-scores from clustering is high ($R^{2}$=$0.62$).

The pairs $c_{13}$-$c_{15}$, $c_{5}$-$c_{13}$ and $c_{5}$-$c_{15}$ that came out on top in Section \ref{sec:winners} also rank highly in terms of 2D-clustering performance.  However, there is hardly any performance gain from using two color components rather than just one.

\begin{figure}[htb]
\begin{center}
\includegraphics[width=0.45\textwidth]{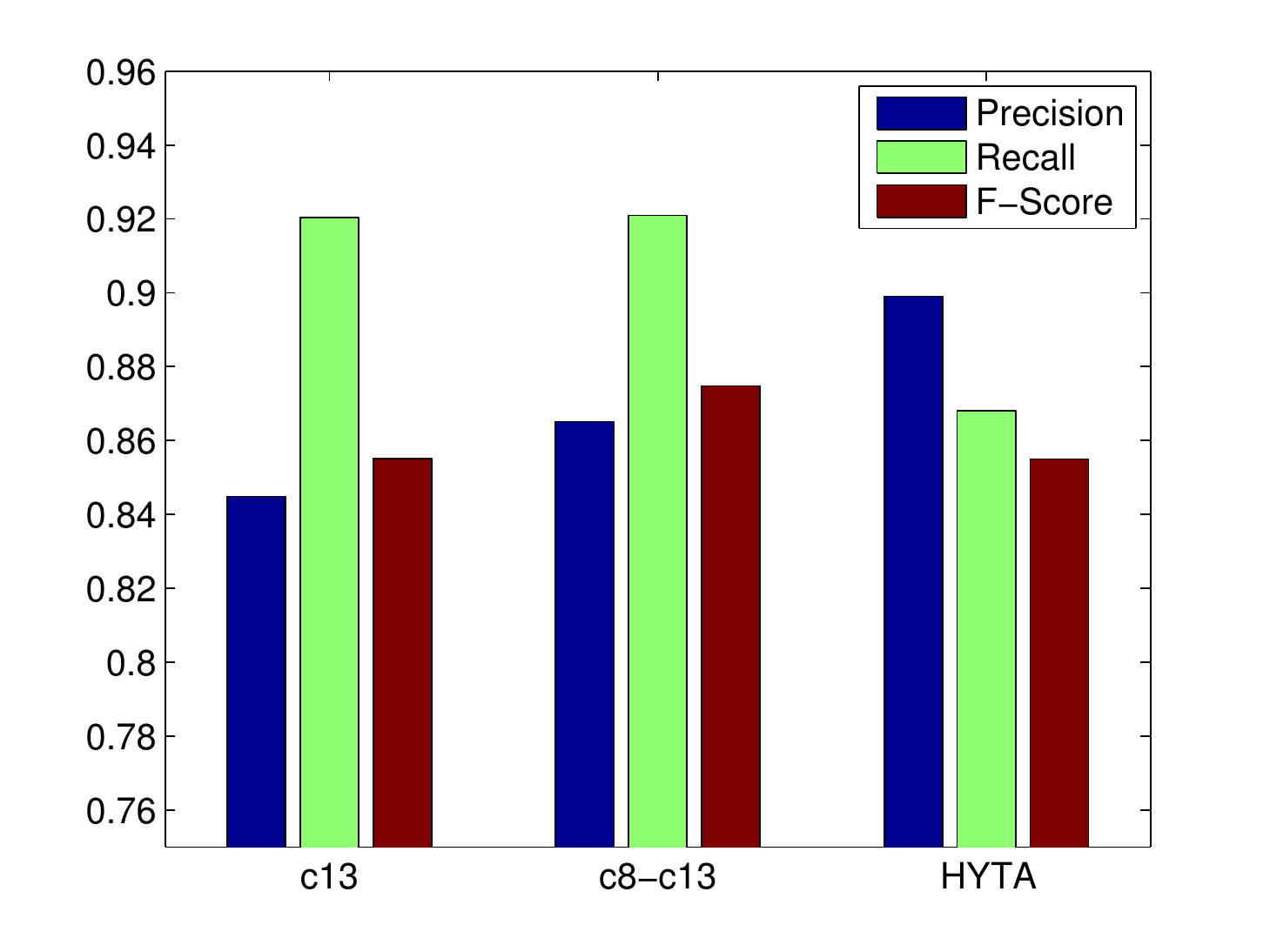}
 \caption{Precision, recall and F-scores for 1D and 2D clustering compared with a state-of-the-art cloud/sky segmentation algorithm \cite{Li2011}.\label{fig:1d_result}}
\end{center}
\end{figure}

As a final comparison, the precision, recall, and F-scores of the best color channels are plotted in Fig.~\ref{fig:1d_result} against the output of a recently proposed cloud detection algorithm \cite{Li2011}.  The data show that a basic 1D clustering method on the most suitable color components and without any tuning or adaptation can achieve a performance that is on par with a much more complex state-of-the-art sky/cloud segmentation method.

\vspace{-0.2in}
\section {Conclusions}
\label{sec:conclusions}
We presented a systematic analysis of color channels for the detection of clouds from sky/cloud images using distribution bimodality, PCA, and clustering.  Experimental evaluation with a cloud segmentation database yields consistent results across analysis methods. Clustering results obtained using the best 1D color channels or 2D color channel pairs (including red-blue combinations $R/B$ and $\frac{B-R}{B+R}$, saturation $S$, or in-phase component $I$) are on par with the performance of a current state-of-the-art cloud detection algorithm in terms of precision, recall, and F-scores.  Future work will involve the additional use of texture and other features to classify clouds into different genera as identified by World Meteorological Organization. 

\balance

\bibliographystyle{IEEEbib}

\begin{thebibliography}{10}

\bibitem{site_diversity}
J.~X. Yeo, Y.~H. Lee, and J.~T. Ong,
\newblock ``Performance of site diversity investigated through {RADAR} derived
  results,''
\newblock {\em IEEE Transactions on Antennas and Propagation}, vol. 59, no. 10,
  pp. 3890--3898, October 2011.

\bibitem{WSI_UCSD}
J.~E. Shields, M.~E. Karr, R.~W. Johnson, and A.~R. Burden,
\newblock ``Day/night whole sky imagers for 24-h cloud and sky assessment:
  history and overview,''
\newblock {\em Applied Optics}, 2013.

\bibitem{Souza}
M.~P. Souza-Echer, E.~B. Pereira, L.~S. Bins, and M.~A.~R. Andrade,
\newblock ``A simple method for the assessment of the cloud cover state in
  high-latitude regions by a ground-based digital camera,''
\newblock {\em Journal of Atmospheric and Oceanic Technology}, vol. 23, no. 3,
  pp. 437--447, March 2006.

\bibitem{Long}
C.~N. Long, J.~M. Sabburg, J.~Calb\'{o}, and D.~Pages,
\newblock ``Retrieving cloud characteristics from ground-based daytime color
  all-sky images,''
\newblock {\em Journal of Atmospheric and Oceanic Technology}, vol. 23, no. 5,
  pp. 633--652, 2006.

\bibitem{WAHRSIS}
S.~Dev, F.~M. Savoy, Y.~H. Lee, and S.~Winkler,
\newblock ``{WAHRSIS}: A low-cost, high-resolution whole sky imager with
  near-infrared capabilities,''
\newblock in {\em Proc. IS\&T/SPIE Infrared Imaging Systems}, 2014.

\bibitem{Li2011}
Q.~Li, W.~Lu, and J.~Yang,
\newblock ``A hybrid thresholding algorithm for cloud detection on ground-based
  color images,''
\newblock {\em Journal of Atmospheric and Oceanic Technology}, vol. 28, no. 10,
  pp. 1286--1296, Oct. 2011.

\bibitem{Calbo2008}
J.~Calb\'{o} and J.~Sabburg,
\newblock ``Feature extraction from whole-sky ground-based images for
  cloud-type recognition,''
\newblock {\em Journal of Atmospheric and Oceanic Technology}, vol. 25, no. 1,
  pp. 3--14, Jan. 2008.

\bibitem{Heinle2010}
A.~Heinle, A.~Macke, and A.~Srivastav,
\newblock ``Automatic cloud classification of whole sky images,''
\newblock {\em Atmospheric Measurement Techniques}, vol. 3, no. 3, pp.
  557--567, 2010.

\bibitem{Sylvio}
S.~L. Mantelli-Neto, A.~von Wangenheim, E.~B. Pereira, and E.~Comunello,
\newblock ``The use of {Euclidean} geometric distance on {RGB} color space for
  the classification of sky and cloud patterns,''
\newblock {\em Journal of Atmospheric and Oceanic Technology}, vol. 27, no. 9,
  pp. 1504--1517, 2010.

\bibitem{pearson}
T.~R. Knapp,
\newblock ``Bimodality revisited,''
\newblock {\em Journal of Modern Applied Statistical Methods}, vol. 6, no. 1,
  pp. 3, 2007.

\bibitem{Bezdek}
J.~C. Bezdek, R.~Ehrlich, and W.~Full,
\newblock ``{FCM}: The fuzzy c-means clustering algorithm,''
\newblock {\em Computers \& Geosciences}, vol. 10, no. 2-3, pp. 191--203, 1984.

\bibitem{Jawahar}
C.~V. Jawahar, P.~K. Biswas, and A.~K. Ray,
\newblock ``Investigations on fuzzy thresholding based on fuzzy clustering,''
\newblock vol. 30, no. 10, pp. 1605--1613, Oct. 1997.

\end{thebibliography}
\end{document}